%% file: paper.tex
\documentclass{article}

\usepackage[preprint]{corl_2022} % Uncomment for pre-prints (e.g., arxiv); This is like ``final'', but will remove the CORL footnote.
\input{header}

\title{Frame Mining: a Free Lunch for Learning Robotic Manipulation from 3D Point Clouds}

\author{
  Minghua Liu\textsuperscript{1*}, Xuanlin Li\textsuperscript{1*}, Zhan Ling\textsuperscript{1*}, Yangyan Li\textsuperscript{2}, Hao Su\textsuperscript{1}\\
  \textsuperscript{1}UC San Diego \quad \textsuperscript{2}Alibaba \\
  \url{https://colin97.github.io/FrameMining/}
}

\begin{document}
\maketitle

%===============================================================================

\blfootnote{
\vspace{-1em}
*equal contribution}

\vspace{-3em}
\input{sections/abstract}
\vspace{-1em}

%===============================================================================

\input{sections/introduction}

\input{sections/related_work}
\input{sections/single_frame}

\input{sections/multi_frame_fusion}

\input{sections/real_world_experiments}

\input{sections/conclusion}

% The maximum paper length is 8 pages excluding references and acknowledgements, and 10 pages including references and acknowledgements

% The acknowledgments are automatically included only in the final and preprint versions of the paper.
\acknowledgments{This work is supported in part by gifts from Qualcomm. We would like to thank Jiayuan Gu and Zhiao Huang for their helpful discussion and manuscript proofreading.}

%===============================================================================

\bibliography{references}  % .bib

\clearpage

\input{sections/supplementary}

\end{document}

%% file: header.tex
\usepackage{graphicx}
\usepackage{wrapfig}
\usepackage{blindtext}
\usepackage{enumitem}
\usepackage{amsfonts}
\usepackage{booktabs}
\usepackage{caption}
\usepackage{subcaption}

\newcommand\blfootnote[1]{%
  \begingroup
  \renewcommand\thefootnote{}\footnote{#1}%
  \addtocounter{footnote}{-1}%
  \endgroup
}

%% file: sections/abstract.tex
\begin{abstract}
 We study how choices of input point cloud coordinate frames impact learning of manipulation skills from 3D point clouds. There exist a variety of coordinate frame choices to normalize captured robot-object-interaction point clouds. We find that different frames have a profound effect on agent learning performance, and the trend is similar across 3D backbone networks. In particular, the end-effector frame and the target-part frame achieve higher training efficiency than the commonly used world frame and robot-base frame in many tasks, intuitively because they provide helpful alignments among point clouds across time steps and thus can simplify visual module learning. Moreover, the well-performing frames vary across tasks, and some tasks may benefit from multiple frame candidates. We thus propose FrameMiners to adaptively select candidate frames and fuse their merits in a task-agnostic manner. Experimentally, FrameMiners achieves on-par or significantly higher performance than the best single-frame version on five fully physical manipulation tasks adapted from ManiSkill and OCRTOC. Without changing existing camera placements or adding extra cameras, point cloud frame mining can serve as a free lunch to improve 3D manipulation learning.
\end{abstract}

%% file: sections/introduction.tex
\vspace{-0.5em}
\section{Introduction}
\vspace{-0.5em}

\begin{wrapfigure}{r}{0.45\textwidth}
    \centering
    \vspace{-2em}
    \includegraphics[width=0.45\textwidth]{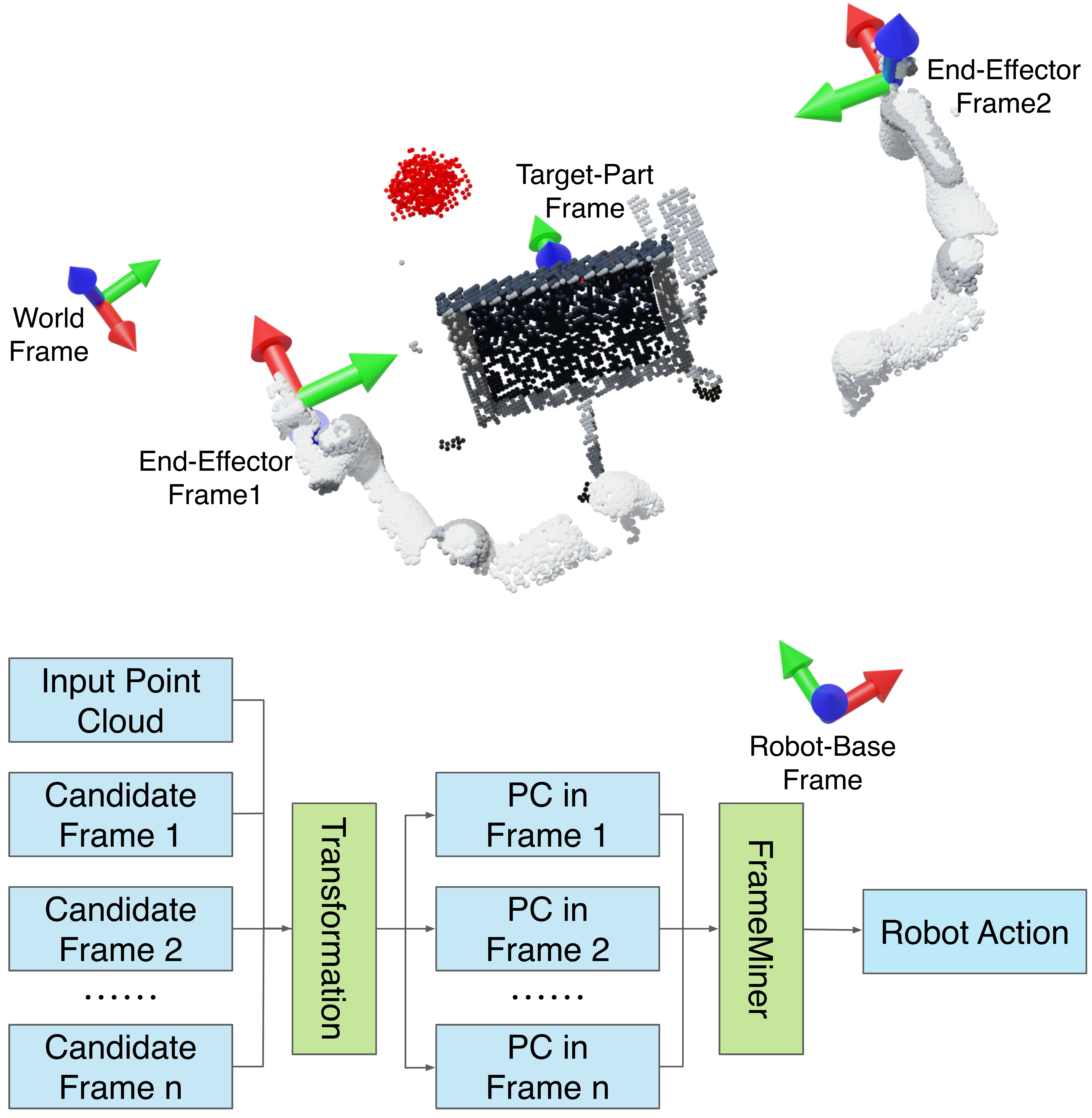}
    \caption{A 3D point cloud of a dual-arm robot pushing a chair, which can be represented in various coordinate frames without changing camera placements or requiring extra camera views. Our FrameMiner takes as input a point cloud represented in multiple candidate frames and adaptively fuses their merits, resulting in better performance.
    }    
    \label{fig:teaser}
    \vspace{-1em}
\end{wrapfigure}

With the rapid development and proliferation of low-cost 3D sensors, point clouds have become more accessible and affordable in robotics tasks~\cite{Kinova}. Also, the tremendous progress in building neural networks with 3D point clouds~\cite{qi2017pointnet,qi2017pointnet++,chang2015shapenet,qi2018frustum,fan2017point,wang2019dynamic} has enabled powerful and flexible frameworks for 3D visual understanding tasks such as 3D object detection~\citep{qi2018frustum,lang2019pointpillars,qi2019deep}, 6D pose estimation~\cite{he2020pvn3d,wang2019densefusion}, and instance segmentation~\cite{yang2019learning,yi2019gspn}. Very recently, point cloud started to be used as the input to deep reinforcement learning (RL) for object manipulation~\citep{huang2021generalization,chen2022system,wu2022learning}, which aims at learning mappings directly from raw 3D sensor observations of unstructured environments to robot action commands. These end-to-end learning methods avoid highly structured pipelines and laborious human engineering required by conventional robot manipulation systems. 

When building an agent with point cloud input, existing works~\cite{huang2021generalization,chen2022system,wu2022learning} typically incorporate off-the-shelf point cloud backbone networks (e.g., PointNet~\cite{qi2017pointnet}) into the pipeline as a feature extractor of the scene. However, some facets in constructing point cloud representations have been overlooked. For example, in the literature of 3D deep learning, the choice of coordinate frame significantly affects task performance~\cite{qi2017pointnet,zhou2020endlin2022learning,alnaggar2021multi,gerdzhev2021tornado,liong2020amvnet,deng2021vector}. On 3D instance segmentation benchmarks for autonomous driving, previous work such as \cite{qi2018frustum} showed a pipeline to process input point clouds in the camera frame, frustum frame, and object frame subsequently, leading to a large performance boost in comparison to using the camera frame alone. For our goal of manipulation skill learning, point clouds describe dynamic interactions between robots and objects, including frequent contacts and occlusions. This is a novel and more complex setting that differs from well-explored scenarios in 3D supervised learning (e.g., single objects, outdoor scenes for autonomous driving). Under this setting, choices of coordinate frames are more flexible and diverse as multiple entities (e.g., robot and manipulated object) and dynamic movements are involved. 

In this work, we first examine whether and how different coordinate frames may impact the performance and sample efficiency of point cloud-based RL for object manipulation tasks. We study four candidate coordinate frames: world frame, robot-base frame, end-effector frame, and target-part frame. These frames differ in positions of origin and orientations of axes, and canonicalize inputs in different manners (e.g., a fixed third-view, ego-centric, hand-centric, object-centric). The comparison and analysis are performed on five distinct physical manipulation tasks adapted from ManiSkill~\cite{mu2021maniskill} and OCRTOC~\cite{liu2021ocrtoc}, covering various numbers of arms, robot mobilities, and camera settings. Results show that the choice of frames has profound effects. In particular, the end-effector frame and the target-part frames, rarely considered in previous works, lead to significantly better sample efficiency and final convergence than the widely used world frame and robot-base frame on many tasks. Visualization and analysis indicate that, by using different coordinate frames to represent input point clouds, we are actually performing various alignments of input scenes through $\mathbb{SE}(3)$ transformations, which may simplify the learning of visual modules. 

However, the well-performing single coordinate frame may vary from task to task, and in many cases, we may need coordination between decisions made according to multiple coordinate frames. For example, tasks equipped with dual-arm robots may benefit from both left-hand and right-hand frames. For mobile manipulation tasks involving both navigation and manipulation, different frames could favor different skills (e.g., robot-base frame for navigation skills, end-effector frame for manipulation skills). We thus propose three task-agnostic strategies to adaptively select from multiple candidate coordinate frames and fuse their merits, leading to more efficient and effective object manipulation policy learning. Because we do not need to capture additional camera views or rely on task-specific frame selections, our frame mining strategies can be used as a free lunch to improve existing methods on point cloud-based policy learning. We call these fusion approaches as \textit{FrameMiners}. Experimentally, we find that it matters to fuse information from multiple frames, but the specific FrameMiner to choose does not create much performance difference. In particular, we use one of the FrameMiners, MixAction, to interpret the importance of different frames in the policy execution process, and the interpretation agrees with our intuitions. 

In summary, the main contributions of this work are as follows:
\vspace{-0.7em}
\begin{itemize}[leftmargin=*]
\setlength\itemsep{0em}
    \item We find that the choice of coordinate frame has a profound impact on point cloud-based object manipulation learning. In particular, the end-effector frame and the target-part frame lead to much better sample efficiency than the widely-used world frame and robot-base frame on many tasks;
    \item We find that well-performing frames differ task by task, necessitating task-agnostic ways to select and fuse frames. This observation is consistent across 3D backbone networks;
    \item We propose FrameMiners, a collection of methods to fuse information from multiple candiate frames. FrameMiners provide a free lunch to improve existing point cloud-based manipulation learning methods without changing camera placements or requiring additional camera views. 
\end{itemize}

%% file: sections/related_work.tex
\vspace{-1em}
\section{Related Work}
\vspace{-1em}

\paragraph{Manipulation Learning with Point Clouds}
Visual representation learning for object manipulation has been extensively studied~\citep{zhu2018reinforcement,chen2021unsupervised,vecerik2020s3k,cheng2018reinforcement,zaky2020active,shah2021rrl,pari2021surprising,nair2022r3m}. With the flourishment of 3D deep learning~\cite{qi2017pointnet,qi2017pointnet++,chang2015shapenet,qi2018frustum,fan2017point,wang2019dynamic}, a major line of work learns representations from 3D point clouds for object manipulation~\cite{lin2022learning,alliegro2022end,yarats2020image,lv2022sagci,james2022q,sundermeyer2021contact,eisner2022flowbot3d}. Recently, people have also started to incorporate point clouds into deep reinforcement learning (RL) pipelines for manipulation learning~\citep{huang2021generalization,chen2022system,wu2022learning}. However, existing point cloud-based manipulation learning methods have not paid enough attention to coordinate frame selections of input point clouds, which is fundamental in 3D visual learning. Some very recent work~\citep{hsu2022vision,jangir2022look} explored placement and selection of camera views and fusion of multi-view images. We differ from them in that we focus on the preprocessing of captured input point clouds without modifying existing camera configurations or adding additional cameras. 

\vspace{-1em}
\paragraph{Normalization and View Fusion in Point Cloud Learning}
Normalizing input point clouds is a common practice in 3D deep learning literature. For example, in single object analysis (e.g., classification and part segmentation), people often normalize input point clouds into a categorical canonical pose with unit scale~\citep{qi2017pointnet,qi2017pointnet++}, simplifying network training. Prior works find that existing point cloud networks~\cite{qi2017pointnet,qi2017pointnet++,wang2019dynamic,atzmon2018point,li2018pointcnn} are very sensitive to input normalization~\cite{deng2021vector,sun2022benchmarking,lorenz2021robustness}, and many recent attempts explore rotation invariant~\cite{chen2019clusternet,li2021rotation,zhang2019rotation} and equivariant methods~\cite{deng2021vector,fuchs2020se,thomas2018tensor} for 3D deep learning. Compared to well-studied scenarios (e.g., single object and autonomous driving), normalizations of point clouds under robot-object interactions are under-explored. 

In LiDAR point cloud learning for autonomous driving, many work focuses on the fusion of multiple views~\cite{zhou2020endlin2022learning,alnaggar2021multi,gerdzhev2021tornado,liong2020amvnet}. Unlike fusing multiple camera scans, there is only one point cloud. They propose to process the point cloud from different views (e.g., perspective view and birds-eye view) to combine their merits, which has proven to be helpful. Our work shares a similar idea. However, we focus on robotic object manipulation settings, and the choice of coordinate systems is more diverse.

%% file: sections/single_frame.tex
\vspace{-1.5em}
\section{Point Cloud Coordinate Frame Selection Matters}
\vspace{-0.8em}

\subsection{Problem Setup}
\vspace{-0.8em}

\begin{figure}[t]
    \centering
    \includegraphics[width=\textwidth]{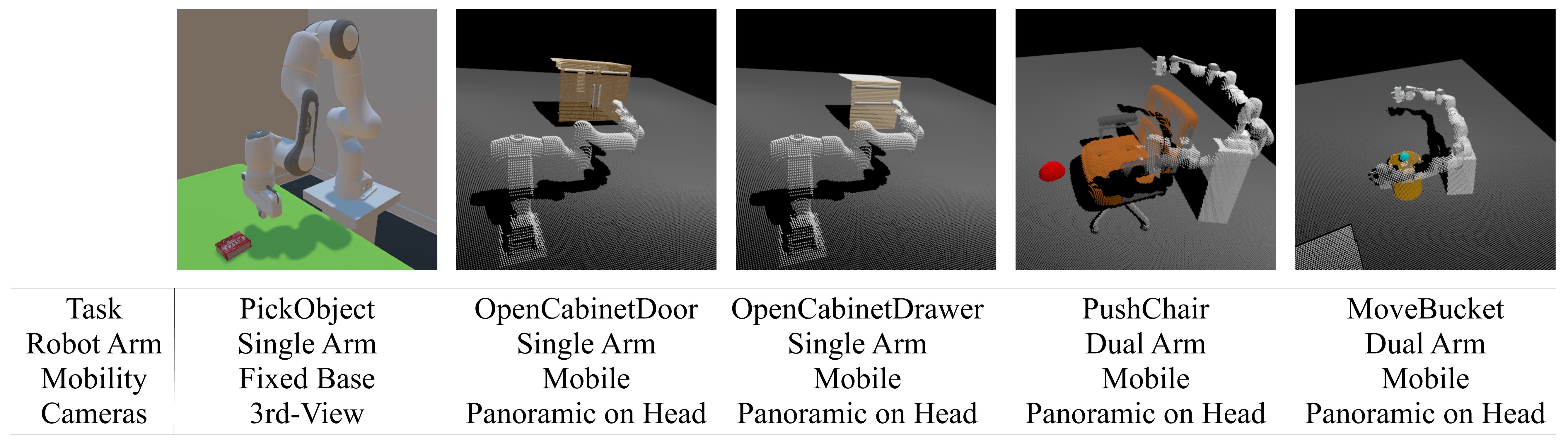}
    \vspace{-1.4em}
    \caption{We study coordinate frame mining on manipulation tasks adapted from OCRTOC~\cite{liu2021ocrtoc} and ManiSkill~\cite{mu2021maniskill} covering various setups (e.g., \#arms, mobility, camera). Simulation is fully physical.}
    \label{fig:tasks}
    \vspace{-0.8em}
\end{figure}

\begin{wrapfigure}{r}{0.6\textwidth}
    \vspace{-1.5em}
    \centering
    \includegraphics[width=0.6\textwidth]{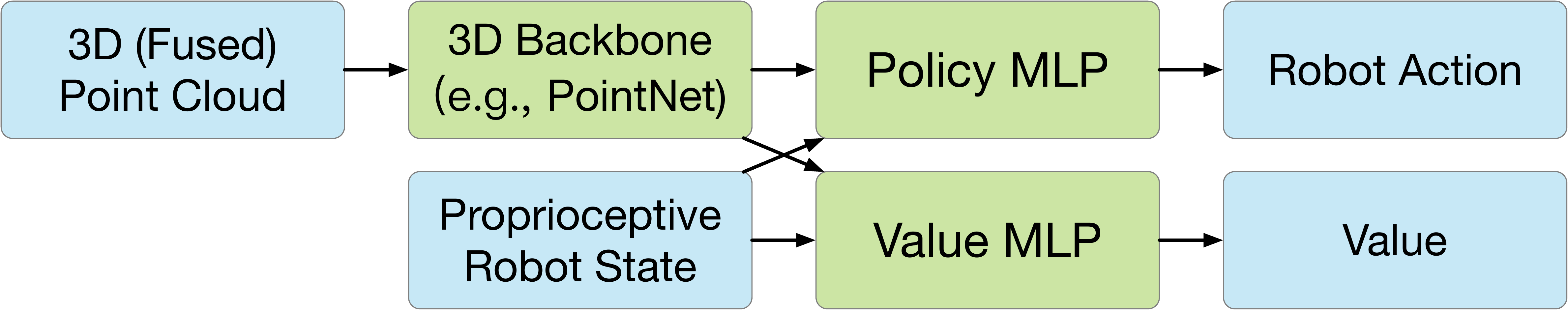}
    \caption{Architecture of a 3D point cloud-based agent, which is optimized by actor-critic RL algorithms. We study coordinate frame selection of input (fused) point cloud.}
    \label{fig:single_frame_pipeline}
    \vspace{-0.5em}
\end{wrapfigure}

We aim to learn agents with point cloud input for object manipulation tasks via Reinforcement / Imitation Learning (RL/IL). A task is formally defined as a Partially-Observable Markov Decision Process (POMDP), which is represented by a tuple $M=(S, A, \mu, T, R, \gamma, \Omega, O)$. Here $S$  and $A$ are the environment state space and the action space. $\mu(s)$, $T(s'|s, a)$, $R(s, a)$, and $\gamma$ are the initial state distribution, state transition probability, reward function, and discount factor, respectively. $O(s): S \to \Omega$ is the observation function that maps environment states to the observation space $\Omega$. Our agent is represented by a policy $\pi: \Omega \to A$, which aims to maximize the expected accumulated return given by $J(\pi \circ O)=E_{\mu, T, \pi}[\sum_{t=0}^{\infty} \gamma^t r(s_t, a_t)]$. Note that $\pi$ does not have access to the environment state $s$ and only has access to the observation $O(s)$. In this work, $O(s)$ consists of two parts: (1) a 3D point cloud captured by depth cameras; (2) proprioceptive states for the robot, such as joint positions and joint velocities. For the first part, if there are multiple cameras, we fuse all point clouds into a single one by transforming them into the same coordinate frame and concatenating the points together. 

Fig.~\ref{fig:single_frame_pipeline} shows the architecture of a 3D point cloud-based agent, which we use to discuss in this section. It first exploits a 3D backbone (e.g., PointNet~\cite{qi2017pointnet}) to extract visual features from a 3D (fused) point cloud. The extracted features are then concatenated with proprioceptive robot states and fed into separate multi-layer perceptrons (MLP) for action and value prediction. The input (fused) point cloud can be represented in different coordinate frames before being fed into the 3D backbone network, and \textit{the choice of coordinate frame is independent of camera views}. For example, a point cloud captured by a camera mounted on the robot's head can be transformed into the end-effector frame. In this work, we study how point cloud coordinate frames affect sample efficiency and final convergence of object manipulation learning. Unlike prior works~\citep{hsu2022vision,jangir2022look}, we do not change robot camera configurations (e.g., camera placement, inclusion of additional cameras).

As shown in Fig.~\ref{fig:tasks}, we exemplify the frame selection problem on five fully-physical manipulation tasks, covering various numbers of robot arms, mobilities, and camera settings. Among them, PickObject is adapted from OCRTOC~\cite{liu2021ocrtoc}, and the other four tasks are adapted from ManiSkill~\cite{mu2021maniskill}. On PickObject, a fixed-base single-arm robot learns to physically grasp an object from the table, lift it up to a target height, and keep it static for a while. Point clouds are captured from a 3rd-view camera. On ManiSkill tasks, agents learn generalizable physical manipulation skills (i.e.,  opening cabinet doors / drawers, pushing chairs / moving buckets to target positions) across objects with diverse topology, geometry, and appearance. We utilize mobile robots with one or two arms. Point clouds come from a panoramic camera mounted on the robot's head. Action space includes joint velocities of the arm(s) and the mobile robot base, along with joint positions of the gripper(s). More details are presented in Appendix~\ref{sec:app_task_descriptions}. 

\vspace{-0.8em}
\subsection{Choices of Point Cloud Coordinate Frame}
\vspace{-0.8em}
\label{sec:choice_single_coord_frame}

\begin{figure}[t]
    \centering
    \includegraphics[width=\textwidth]{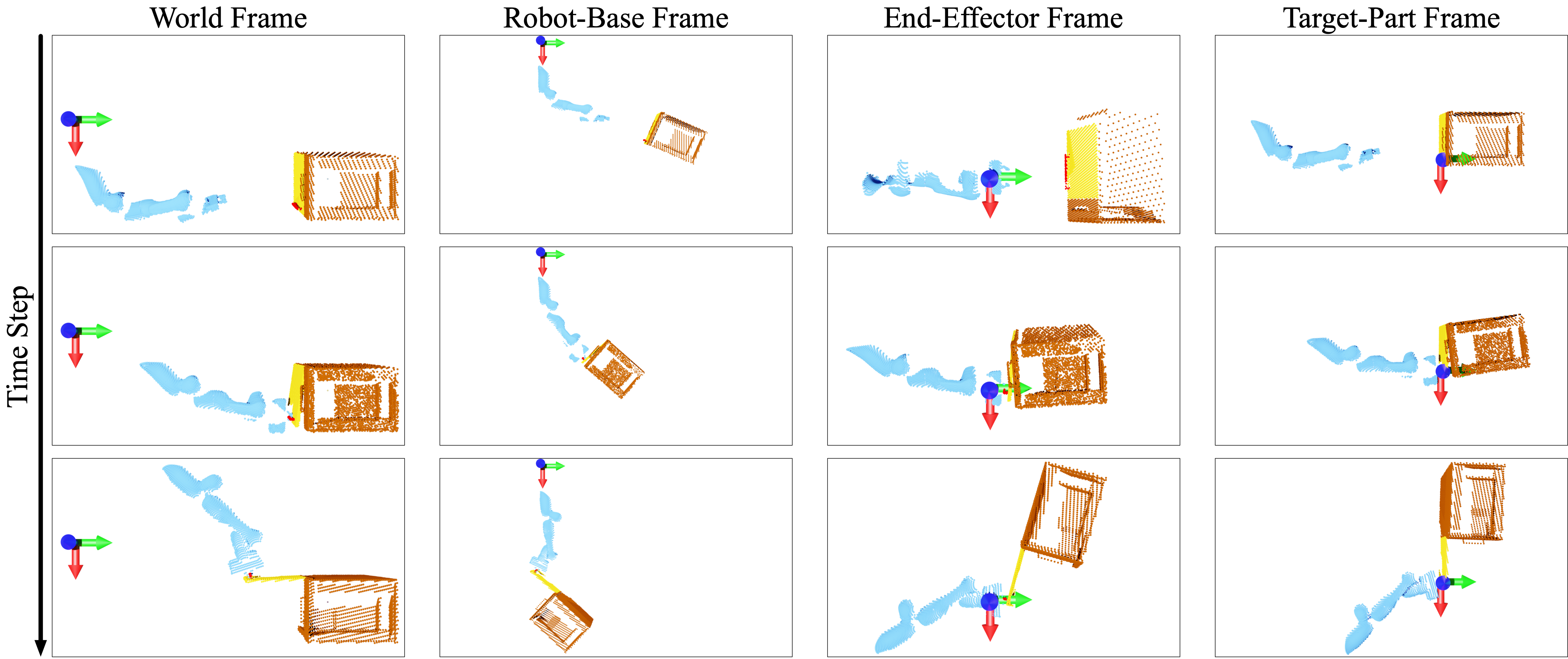}
    \vspace{-1.0em}
    \caption{Illustration of four coordinate frames, which provide different alignments across time steps. We visualize three point clouds (three time steps) of an OpenCabinetDoor trajectory. Each row shows the same point cloud represented in different coordinate frames. Please zoom in for details. Robot arm, cabinet door handle, cabinet door, and cabinet body are colored in blue, red, yellow, and brown, respectively. RGB arrows indicate the corresponding origin and axes for each frame. Since the point clouds used for policy learning can be rather sparse, we show dense point clouds here for better visualization.}
    \label{fig:four_frames}
    \vspace{-0.5em}
\end{figure}

For 3D supervised learning tasks such as object classification and detection, it's a common practice to normalize input point clouds, and the choice of coordinate frames significantly affects task performance~\citep{qi2017pointnet,deng2021vector,zhou2020endlin2022learning,alnaggar2021multi,gerdzhev2021tornado,liong2020amvnet}. In point cloud-based manipulation learning, we are faced with an underexplored, yet more challenging, setting. First, point clouds describe more complex robot-object interactions, possibly including frequent contacts and occlusions. Furthermore, compared to supervised learning, 3D visual modules receive weaker supervision signals during RL training. Therefore, it may become even more important to lessen the burden of visual module learning by properly normalizing input point clouds. Unlike previous well-studied point cloud learning scenarios (e.g., single-object point clouds, LiDAR point clouds for autonomous driving), there exist more diverse choices of coordinate frames. In this paper, we compare and analyze four candidates: 
\vspace{-0.7em}
\begin{itemize}[leftmargin=*]
\setlength\itemsep{0em}
    \item  A \textbf{world frame} is attached to a fixed point in the world (e.g., the start point of a trajectory).
    \item A \textbf{robot-base frame} is attached to the robot base, offering an egocentric perspective on a mobile robot. For a fixed-base robot, world frame and robot-base frame could be equivalent.
    \item In many object manipulation tasks, movements of robot end-effector(s) play important roles, and we can attach an \textbf{end-effector frame} to each of them. Note that for dual-arm robots, there are two end-effectors and thus two end-effector frames.
    \item A \textbf{target-part frame} is attached to the object part the robot intends to interact with (e.g., target door handle for the OpenCabinetDoor task).
\end{itemize}
\vspace{-0.7em}

When we transform captured point clouds into the world frame, the robot-base frame, and the end-effector frame, we may need proprioceptive robot states and potential robot movement tracking, which is typically accessible in modern robots. When we transform point clouds into the target-part frame, we may need to leverage off-the-shelf 3D object detection and pose estimation techniques. However, in this paper, we mainly focus on the choices of coordinate frames themselves. In our simulated experiments, we use ground truth object poses for the target-part frame.

In Fig.~\ref{fig:four_frames}, we visualize an example trajectory under four coordinate frames. As shown in the figure, different coordinate frames canonicalize inputs in different manners (e.g., a fixed third-view, ego-centric, hand-centric, and object-centric), which is essentially performing various \textit{alignments} among point clouds across multiple time steps. For example, in the end-effector frame, the end-effector is always aligned at the origin throughout a trajectory. Such alignments may simplify the learning of visual modules in distinct ways. With the end-effector frame, the network does not need to locate the end-effector in point clouds (always at the origin). Similarly, with the target-part frame, it can be easier to determine the relative position between the target part and the robot end-effector. The robot-base frame naturally aligns its frame axes with the moving directions of the robot's base.

\vspace{-0.8em}
\subsection{Single-Frame Comparison on Manipulation Tasks}
\vspace{-0.8em}
\label{sec:single_results}
\begin{figure}[t]
    \centering
    \includegraphics[width=\textwidth]{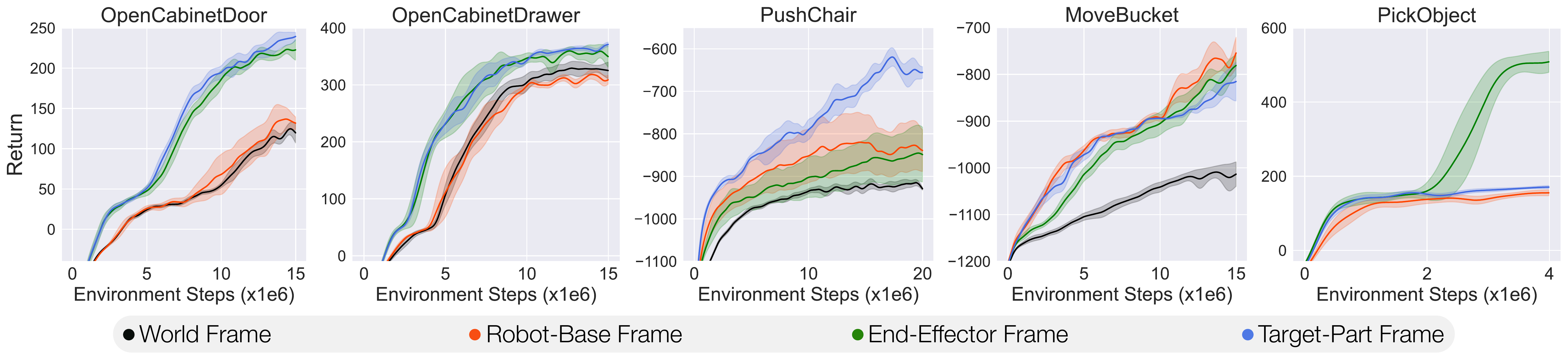}
    \caption{Comparison of four coordinate frames on five fully-physical manipulation tasks. The (fused) point cloud is transformed to a \textit{single} coordinate frame before being fed to the visual backbone network. For dual-arm tasks (i.e., PushChair and MoveBucket), we use the right-hand frame as the end-effector frame. For PickObject, which has a fixed base, the world frame is the same as the robot-base frame. Mean and standard deviation over 5 seeds are shown.}
    \vspace{-0.5em}
    \label{fig:single_frame_comparison}
\end{figure}

We compare the four coordinate frames on the five manipulation tasks by training PPO~\cite{schulman2017proximal} agents using PointNet~\cite{qi2017pointnet} as the 3D visual backbone. In this section, the (fused) point cloud is transformed into a \textbf{single} coordinate frame. For PushChair and MoveBucket tasks that use a dual-arm robot, we use the right hand frame as the end-effector frame (we observe almost identical performance using the left hand frame). For the target-part frame, we choose the handle frame for OpenCabinetDoor and OpenCabinetDrawer tasks, chair seat frame for the PushChair task, bucket for the MoveBucket task, and the target object for the PickObject task. Further details are presented in the supplementary. 

Fig.~\ref{fig:single_frame_comparison} shows the results. We observe that distinct coordinate frames lead to very different agent training performances. Overall, the world frame is the least effective, especially in PushChair and MoveBucket that involve more pronounced movement of the robot base. This suggests that the alignment of a static point in the world-frame is less helpful for the tasks. Compared to the commonly-used world-frame and robot-base frame, the end-effector frame has much higher sample efficiency on all single-arm tasks (i.e., OpenCabinetDoor, OpenCabinetDrawer, and PickObject), demonstrating the benefits of end-effector alignment. However, it shows similar or worse performance on PushChair and MoveBucket, intuitively because these tasks rely on dual-arm coordination, but our point cloud is normalized to a single end-effector frame (i.e., right hand frame). In addition, the target-part frame achieves the best sample efficiency on most tasks, suggesting that the target-part alignment across time could be of great help for point cloud-based visual manipulation learning.

\vspace{-0.8em}
\subsection{Further Analysis}
\vspace{-0.8em}

\begin{wrapfigure}{r}{0.36\textwidth}
    \centering
    \vspace{-1em}
    \includegraphics[width=0.36\textwidth]{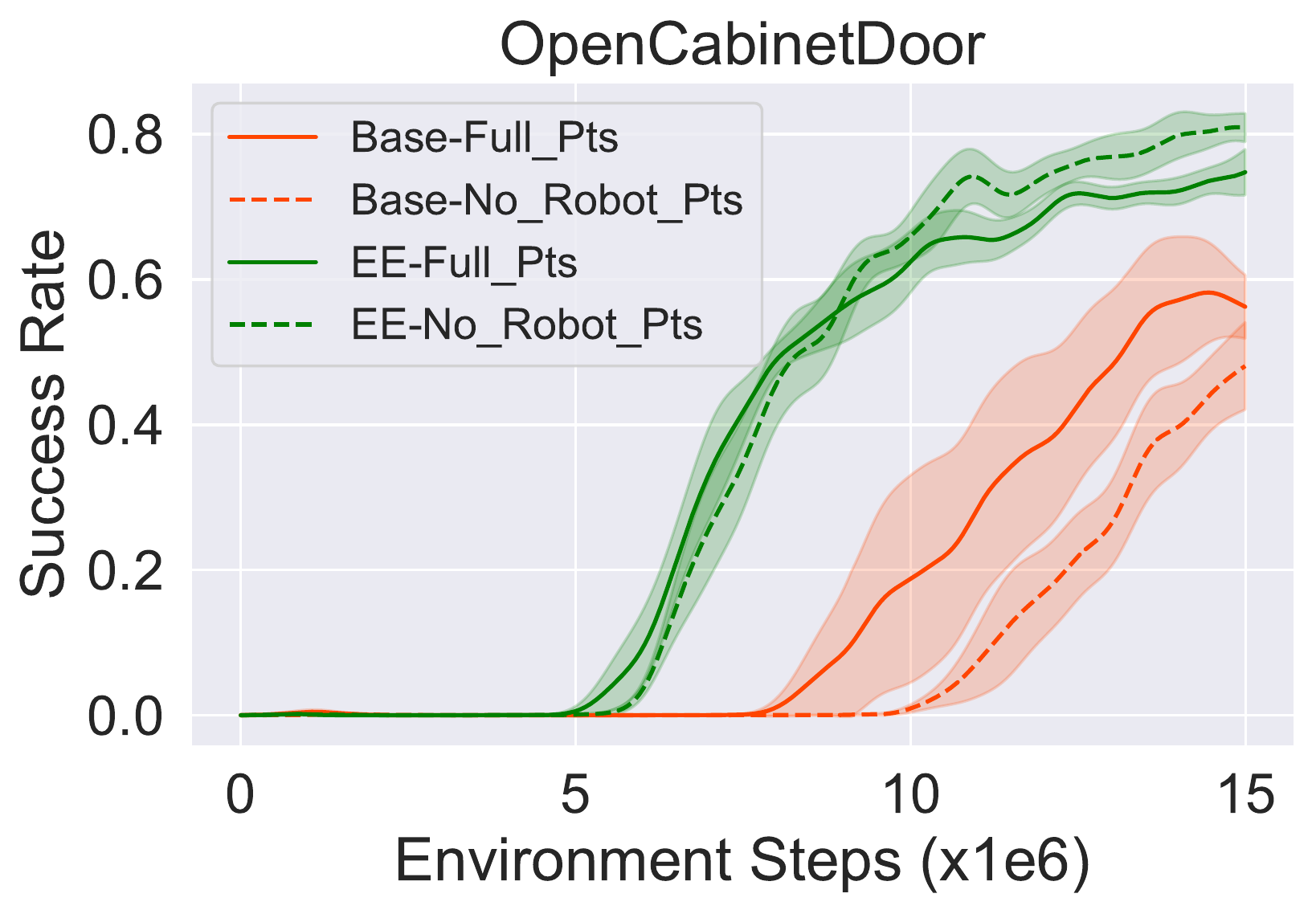}
    % \caption{Performance of the robot-base frame and the end-effector frame, before and after the robot points are removed, on OpenCabinetDoor.}    
    \caption{Removing robot points does not harm the performance of the EE frame on OpenCabinetDoor.}    
    %\caption{On OpenCabinetDoor, the performance of the end-effector frame is not harmed after removing the robot points.}
    \label{fig:no_robot_pts}
    \vspace{-1em}
\end{wrapfigure}

In robot manipulation tasks, agents often need to infer \textit{binary relations} between subjects (e.g., relative pose between the end-effector and the cabinet handle). By aligning point clouds under certain frames (e.g., end-effector frame), these tasks may be reduced to \textit{single-subject location tasks} (e.g., simply copying the handle pose), which become much easier to solve. To confirm this hypothesis, we perform a diagnosis experiment on OpenCabinetDoor, where we intentionally remove all robot points (i.e., blue points in Fig.~\ref{fig:four_frames}) and see its effect on different coordinate frames. As shown in Fig.~\ref{fig:no_robot_pts}, after the robot points are removed, the end-effector frame performs the same, while the robot-base frame performs worse (the task is still solvable since the end-effector position is also provided in the proprioceptive robot state). This suggests that the end-effector frame allows an agent to focus on the target object, along with its interaction with the robot hand, which verifies our hypothesis. 

We utilized PointNet~\cite{qi2017pointnet} as our 3D visual backbone for its fast speed and general good performance. However, it's unclear whether point cloud frame selection is also crucial for other 3D backbones, especially those more complex and powerful. Therefore, we conduct the same experiments as Sec~\ref{sec:single_results} using SparseConvNet~\cite{SubmanifoldSparseConvNet}, a heavier 3D backbone network, on the four ManiSkill tasks (further details in the supplementary). As shown in Fig.~\ref{fig:sparseconv}, we observe similar relative performance between frames as before (e.g., the world frame performs poorly; the end-effector frame outperforms the world frame and the robot-base frame on OpenCabinetDoor and OpenCabinetDrawer). Interestingly, using SparseConvNet doesn't improve the overall performance over PointNet. 

\begin{figure*}[t]
    \centering
    \includegraphics[width=\textwidth]{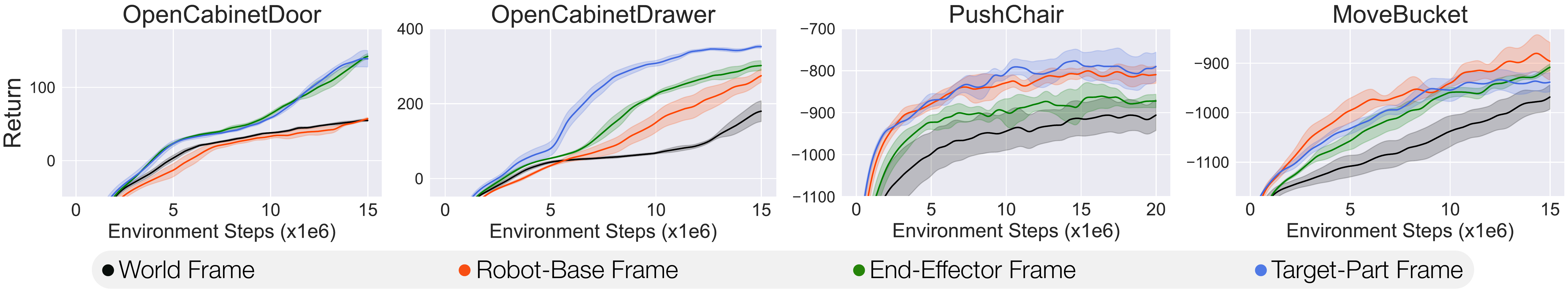}
    \vspace{-1.4em}
    \caption{Using SparseConvNet~\cite{SubmanifoldSparseConvNet} as the 3D visual backbone, we observe similar trends as Fig~\ref{fig:single_frame_comparison}. Mean and standard deviation over 5 seeds are shown.}
    \label{fig:sparseconv}
    \vspace{-0.5em}
\end{figure*}

%% file: sections/multi_frame_fusion.tex
\vspace{-0.5em}
\section{Mining Multiple Coordinate Frames}
\vspace{-0.5em}

\begin{wrapfigure}{r}{0.41\textwidth}
    \vspace{-0.5em}
    \centering
    \includegraphics[width=0.41\textwidth]{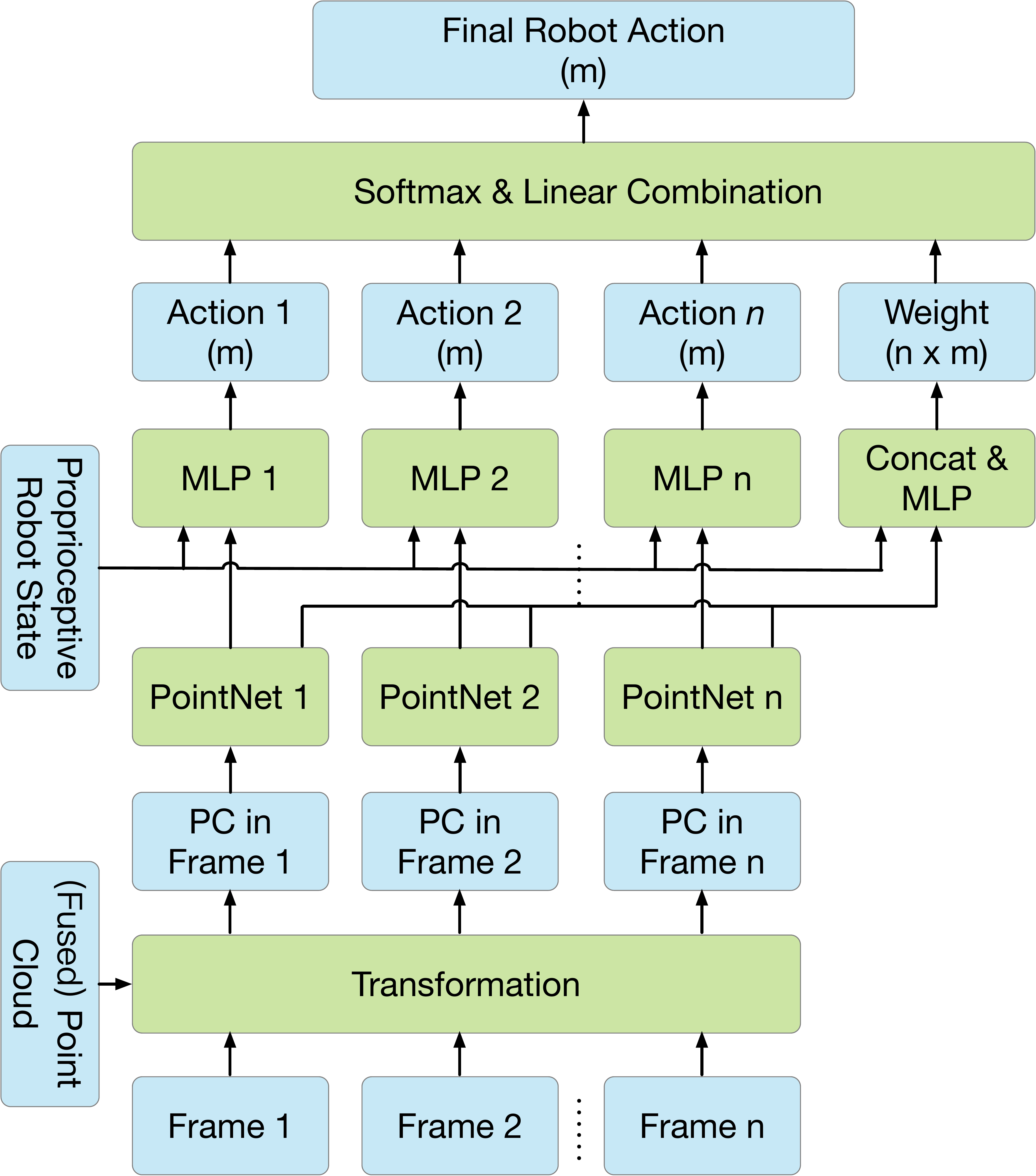}
    \caption{The pipeline of FrameMiner-MixAction (FM-MA). Each frame outputs an action proposal. Actions are then fused through input-dependent and joint-specific weights. }
    \label{fig:frame_miner}
    \vspace{-1.4em}
\end{wrapfigure}	
We have shown that different point cloud coordinate frames lead to distinct sample efficiencies and final performances in manipulation learning. However, a single frame can perform well on some tasks but poorly on others, and we wish to find good frames in a task-agnostic manner. Moreover, for complex manipulation tasks, a single frame could be insufficient, and synergistic coordination between multiple frames could provide unparalleled advantages. For example, when robots are equipped with multiple arms, each arm may have its preferred coordinate frame (e.g., left-hand frame and right-hand frame). In addition, in tasks that involve simultaneous navigation and manipulation (e.g., on PushChair and MoveBucket, an agent needs to move towards the target while manipulating chairs or buckets), different frames could benefit different skills (e.g., robot-base frame for navigation skills, and end-effector frame for manipulation skills). Therefore, it is of great help to propose a prior-agnostic method that can automatically select the best frame from multiple candidates or combine the merits of them. Again, we are not talking about fusing multiple camera views. Point clouds from multiple camera views are first fused together into a single point cloud, before being transformed to each coordinate frame.

In this section, we will present a collection of three strategies to adaptively select and fuse multiple candidate coordinate frames, and we call them \textit{FrameMiners}. In particular, we will first introduce \textit{FrameMiner-MixAction} in Section~\ref{sec:mixaction} in detail to interpret the importance of different frames in the policy execution process. We will then briefly introduce the other two FrameMiners and compare different approaches with single-frame baselines.

\vspace{-0.7em}
\subsection{FrameMiner-MixAction}
\label{sec:mixaction}
\vspace{-0.7em}

Inspired by the idea of mixture of experts~\cite{masoudnia2014mixture}, we propose a general and interpretable framework, \textit{FrameMiner-MixAction} (FM-MA). As shown in Fig.~\ref{fig:frame_miner}, FM-MA takes a (fused) point cloud and $n$ candidate coordinate frames as input, and first transforms the point cloud into $n$ coordinate frames. For each transformed point cloud, FM-MA employs an expert network, consisting of a 3D visual backbone (e.g., PointNet~\cite{qi2017pointnet}) and an MLP, to propose full robot actions (e.g. target velocity of $m$ joints). Since different experts use different coordinate frames, they are encouraged to specialize different skills and controls of different joints. Finally, we combine actions from the $n$ experts through input-dependent weights. Specifically, we concatenate extracted visual features from all $n$ frames with the proprioceptive robot state, feed it into an MLP, and predict a weight for each pair of expert and joint (there are $n\times m$ weights in total). For each joint, we normalize the weights over $n$ experts via softmax and fuse the actions through weighted linear combination. 

\begin{figure*}[t]
    \centering
    \includegraphics[width=\textwidth]{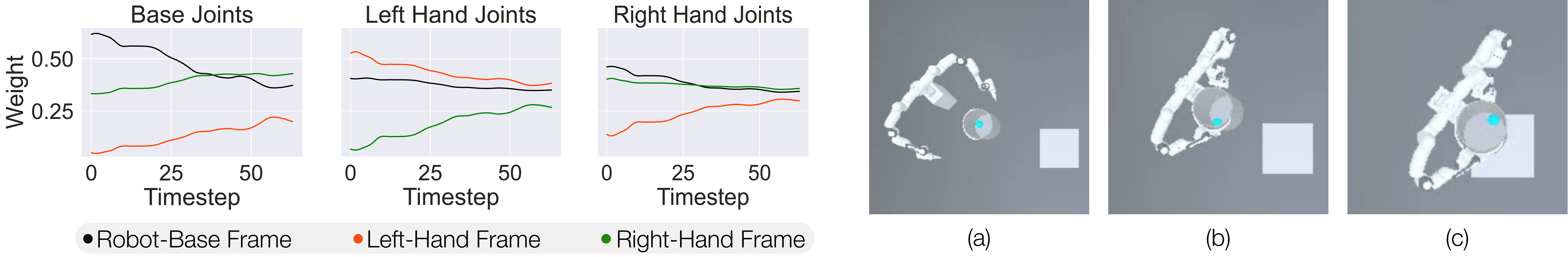}
    \caption{Left: learned weights of FrameMiner-MixAction over a MoveBucket trajectory, where three coordinate frames are fused. We divide robot joints into three groups and show the average weights of each group in each coordinate frame. Right: three stages of the trajectory. (a) Approaching the bucket. (b) Moving the bucket to the platform. (c) Placing the bucket on the platform.}
    \vspace{-1.1em}
    \label{fig:bucket_weights}
\end{figure*}

FM-MA fuses actions by predicting joint-specific weights, since we believe that, for different joints, we need to extract information from different coordinate frames. Furthermore, the weights are input-dependent, potentially allowing the model to capture dynamic joint-frame relations at different stages of a task. Fig.~\ref{fig:bucket_weights} confirms these hypotheses. On MoveBucket trajectories, we observe distinct frame preferences between different robot joints. The left and right-hand frames contribute significantly to their respective joint actions. In addition, the weight distribution changes greatly over different trajectory stages. Initially, when the robot is moving towards the bucket but not interacting with it, the base frame contributes more. However, when the hands start to manipulate the bucket, the hand frames' weights increase. In particular, when the robot places the bucket onto the platform, we need careful coordination among all joints, and thus similar weights from each frame.

\vspace{-1em}
\subsection{FrameMiners vs. Single Coordinate Frame}
\label{sec:multi_fusion_comparison}
\vspace{-0.6em}

To study how network architectures influence coordinate frame fusion, we also propose other two strategies: \textit{FrameMiner-FeatureConcat} (FM-FC) and \textit{FrameMiner-TransformerGroup} (FM-TG). For each transformed point cloud, FM-FC uses an individual PointNet to extract visual feature. All visual features are then concatenated and fed into an MLP to predict robot action. FM-TG decomposes our robot action into three groups: base joint actions, left-hand joint actions, and right-hand joint actions (only two groups for single-arm tasks). After visual features are extracted from PointNets, they are fused through a Transformer~\citep{vaswani2017attention} to produce a feature for each action group, which passes through an MLP to predict its respective joint actions (see Appendix~\ref{sec:app_more_frameminers} for details).

\begin{figure*}[t]
    \centering
    \includegraphics[width=\textwidth]{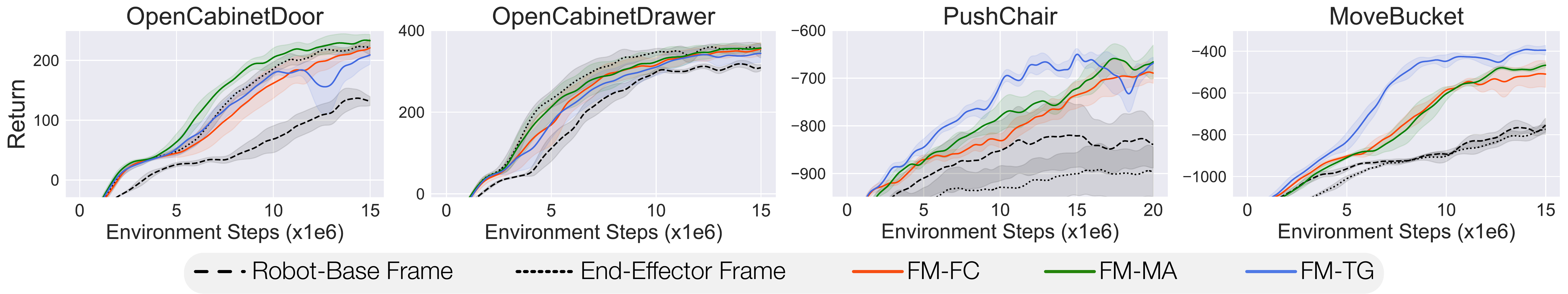}
    \vspace{-1.5em}
    \caption{Comparison of different frame mining approaches on the four ManiSkill tasks, where the robot-base frame and end-effector frame(s) are fused. Black lines indicate single-frame baselines. Mean and standard deviation over 5 seeds are shown.}
    \label{fig:multi_frame_comparison}
    \vspace{-0.5em}
\end{figure*}

\input{tables/multi_frame_comparison_success_rate}
We compare our three FrameMiners with single-frame baselines. \textit{Specifically, in this section, we focus on frame mining among the robot-base frame and end-effector frame(s).} For the dual-arm tasks (i.e., PushChair and MoveBucket), the end-effector frames include both the left-hand frame and the right-hand frame. We will discuss the inclusion of target-part frame in Section~\ref{sec:obj_frame}. Fig.~\ref{fig:multi_frame_comparison} and Tab.~\ref{tab:success_rate} show the comparison results. On single-arm tasks (i.e., OpenCabinetDoor/Drawer), our FrameMiners perform on par with the end-effector frame, which suggests that FrameMiners can automatically select the best single frame. On dual-arm tasks, our FrameMiners significantly outperform single-frame baselines, demonstrating the advantage of coordination between multiple coordinate frames. We also demonstrate that our FrameMiners outperform alternative designs and provide robust advantages (more details in Appendix~\ref{sec:app_frameminer_max_weight},~\ref{sec:app_cam_placement}, and~\ref{sec:app_adaptive_se3}). While it matters to fuse information from multiple frames, the specific FrameMiner to choose does not create much performance difference. Empirically, we find FM-MA less sensitive to training parameters, and FM-TG more computationally expensive.

Our previous experiments were conducted using online RL. To investigate whether our previous findings can generalize to other algorithm domains, we perform experiments on imitation learning, and more details are presented in Appendix \ref{sec:app_imitation}. The results corroborate our previous findings, i.e., different coordinate frames have a profound effect on point cloud-based manipulation skill learning, and FrameMiners are capable of automatically selecting the best coordinate frame or combining the
merits from multiple frames and outperforming single-frame results.

\vspace{-0.8em}
\subsection{Target-Part Frame}
\label{sec:obj_frame}
\vspace{-0.8em}
\begin{wrapfigure}{r}{0.4\textwidth}
    \vspace{-2em}
    \centering
    \includegraphics[width=0.4\textwidth]{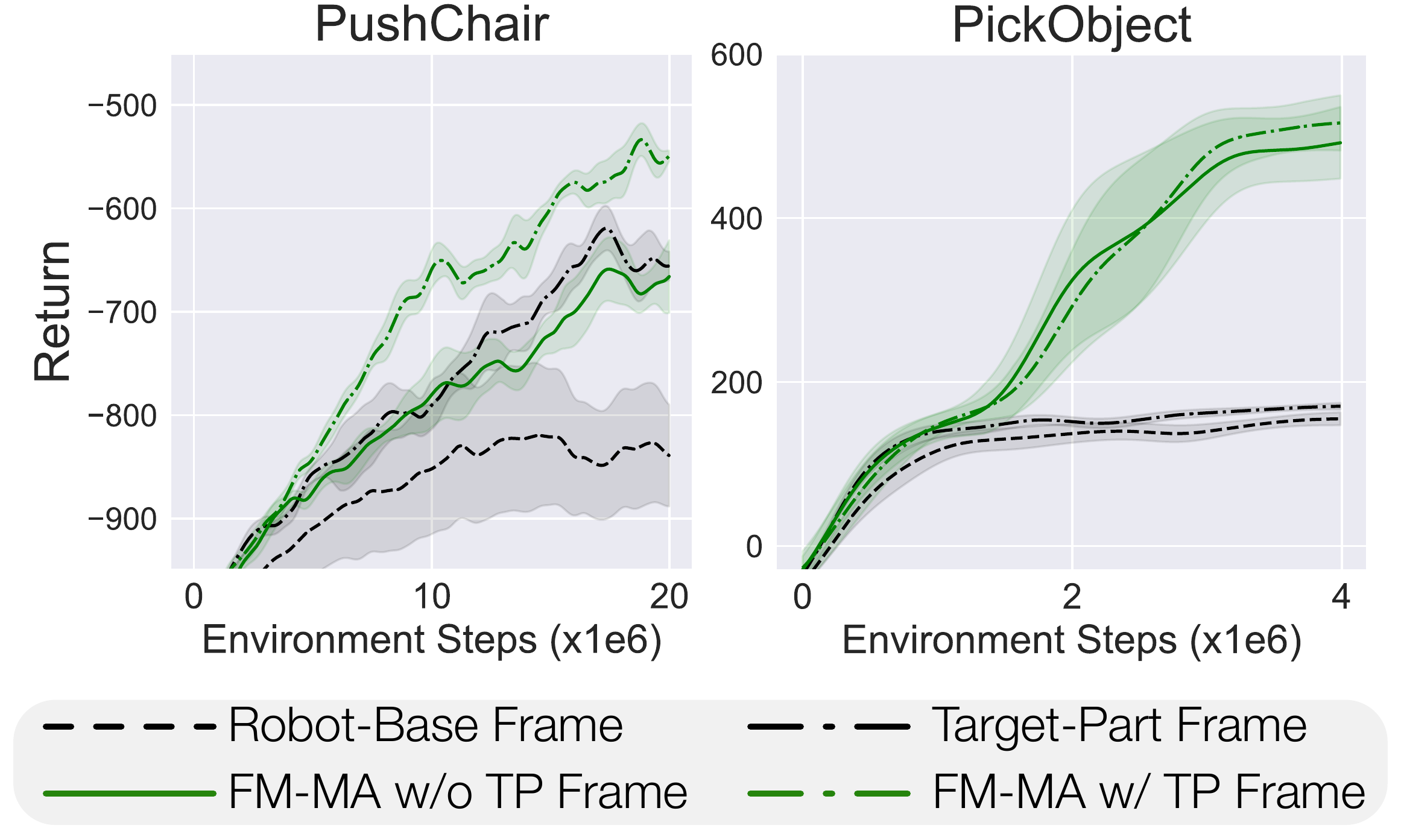}
    \vspace{-1.5em}
    \caption{Fusion of target-part frame could further boost the performance.}
    \vspace{-1em}
    \label{fig:ablation_target_part}
\end{wrapfigure}	
In Section~\ref{sec:multi_fusion_comparison}, we focus on the fusion of robot-base frame and end-effector frames, since the target-part frame relies on pose estimation of target objects, which requires extra efforts in real-world settings. As shown in Fig.~\ref{fig:ablation_target_part}, if object poses are estimated, we sometimes observe further performance boost of FrameMiners from the target-part frame. For example, on PushChair, by incorporating the target-part frame, the success rate of FM-MA increases from 36$\pm$4\% to 53$\pm$3\%. On PickObject, FM-MA already achieves good performance without the target-part frame (94$\pm$2\%); incorporating it slightly improves the success rate to 97$\pm$1\%.

%% file: tables/multi_frame_comparison_success_rate.tex
\begin{wraptable}{r}{0.35\textwidth}
  \scriptsize
  \setlength{\tabcolsep}{1.4pt}
  \vspace{-1.4em}
    \begin{tabular}{c|cc|ccc}
    \toprule
          & Base & EE  & FM-FC & FM-MA & FM-TG \\
    \midrule
    Door &  54$\pm$7    &  80$\pm$2     &  79$\pm$3     &   \textbf{84$\pm$2}    &  70$\pm$6 \\
    Drawer &   88$\pm$2    &  93$\pm$1     &  \textbf{94$\pm$1}     &   93$\pm$1    &  93$\pm$2  \\
    Chair &  7$\pm$3     &  2$\pm$1     &  32$\pm$4     &  \textbf{36$\pm$4}     & 34$\pm$6 \\
    Bucket &   23$\pm$6    &   19$\pm$4    &  77$\pm$5     & 81$\pm$3      & \textbf{90$\pm$2} \\
    \bottomrule
    \end{tabular}%
   \caption{Success rates (\%) on four ManiSkill tasks.}
    \label{tab:success_rate}
    \vspace{-1em}
    
  \label{tab:addlabel}%

\end{wraptable}

%% file: sections/real_world_experiments.tex
\vspace{-1em}
\section{Real World Experiments}
\vspace{-1em}

\begin{wrapfigure}{r}{0.28\textwidth}
    \centering
    \vspace{-1em}
    \includegraphics[width=0.28\textwidth]{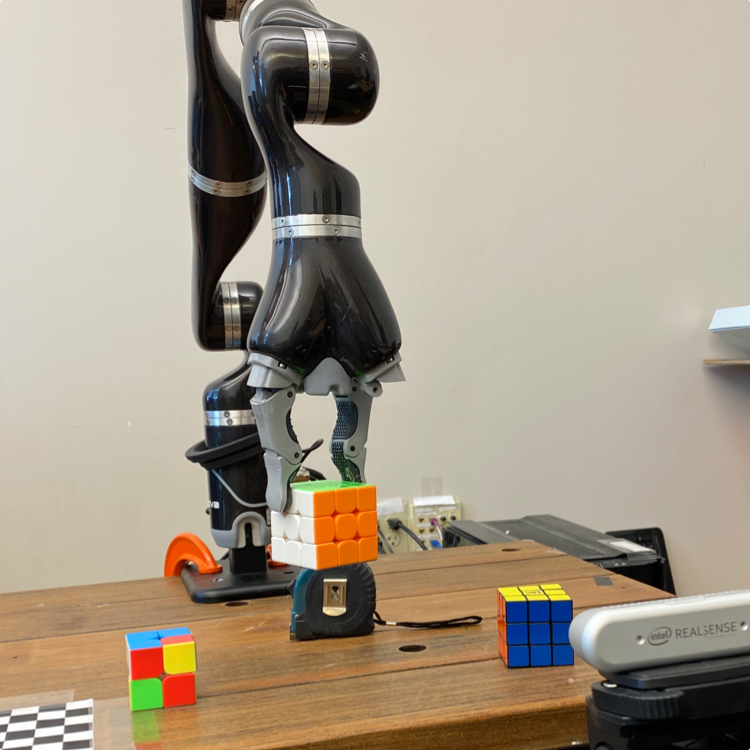}
    \caption{Real robot setup.}
    \vspace{-1em}
    \label{fig:real_robot}
\end{wrapfigure}	

To further verify that our learned policies can be deployed on real-world robots without introducing extra domain gaps, we test on PickObject with a Kinova Jaco2 Spherical 7-DoF robot, an Intel RealSense~\citep{keselman2017intel} D435 camera for uncolored point cloud capture, and a Rubik's cube from YCB objects~\citep{calli2015benchmarking, calli2015ycb} (see Fig. \ref{fig:real_robot}). We use a 3-DoF end-effector position controller and a 1-DoF gripper position controller. We train the FM-MA policy (Section~\ref{sec:mixaction}) by fusing the robot-base frame and the end-effector frame. At test time, we build a digital twin in the simulator over 25 sampled initializations of the real environment with a vision-based pipeline like~\citet{jiang2022ditto}. By following trajectories from policy rollout, we obtain a 84\% success rate with FM-MA, compared to an 80\% success rate with the end-effector frame. The robot-base frame is unable to achieve successful picks under our training budget. Note that the performance differences in the real world are very similar to the simulation environment, indicating that point cloud frame selection or mining does not affect original domain gap. More details are presented in Appendix~\ref{sec:app_real_world}.

%% file: sections/conclusion.tex
\vspace{-1em}
\section{Conclusion and Limitations}
\vspace{-1em}

We find that choices of point cloud coordinate frames have a profound impact on learning manipulation skills. Our proposed FrameMiners can adaptively select and fuse multiple candidate frames, serving as a free lunch for 3D point cloud-based manipulation learning. 
Currently, our FrameMiners need to process each frame separately, leading to more network computation. In the future, we would like to explore more advanced fusion strategies to further improve network efficiency as well as performance. In addition, the target-part frame requires human judgment to determine the target part candidates and 6D pose estimation of the target parts, although we have shown our method can also achieve great improvements without the target-part frame (Section~\ref{sec:multi_fusion_comparison}).

%

%% file: sections/supplementary.tex
\renewcommand{\thesection}{S}

\subsection{Architecture of the other two FrameMiners}
\label{sec:app_more_frameminers}
Fig.~\ref{fig:FMFC,FMTG} shows architectures of the other two FrameMiners, FrameMiner-FeatureConcat (FM-FC) and FrameMiner-TransformerGroup (FM-TG). 

\begin{figure}[t]
     \centering
     \begin{subfigure}[t]{0.42\textwidth}
         \centering
         \includegraphics[width=\textwidth]{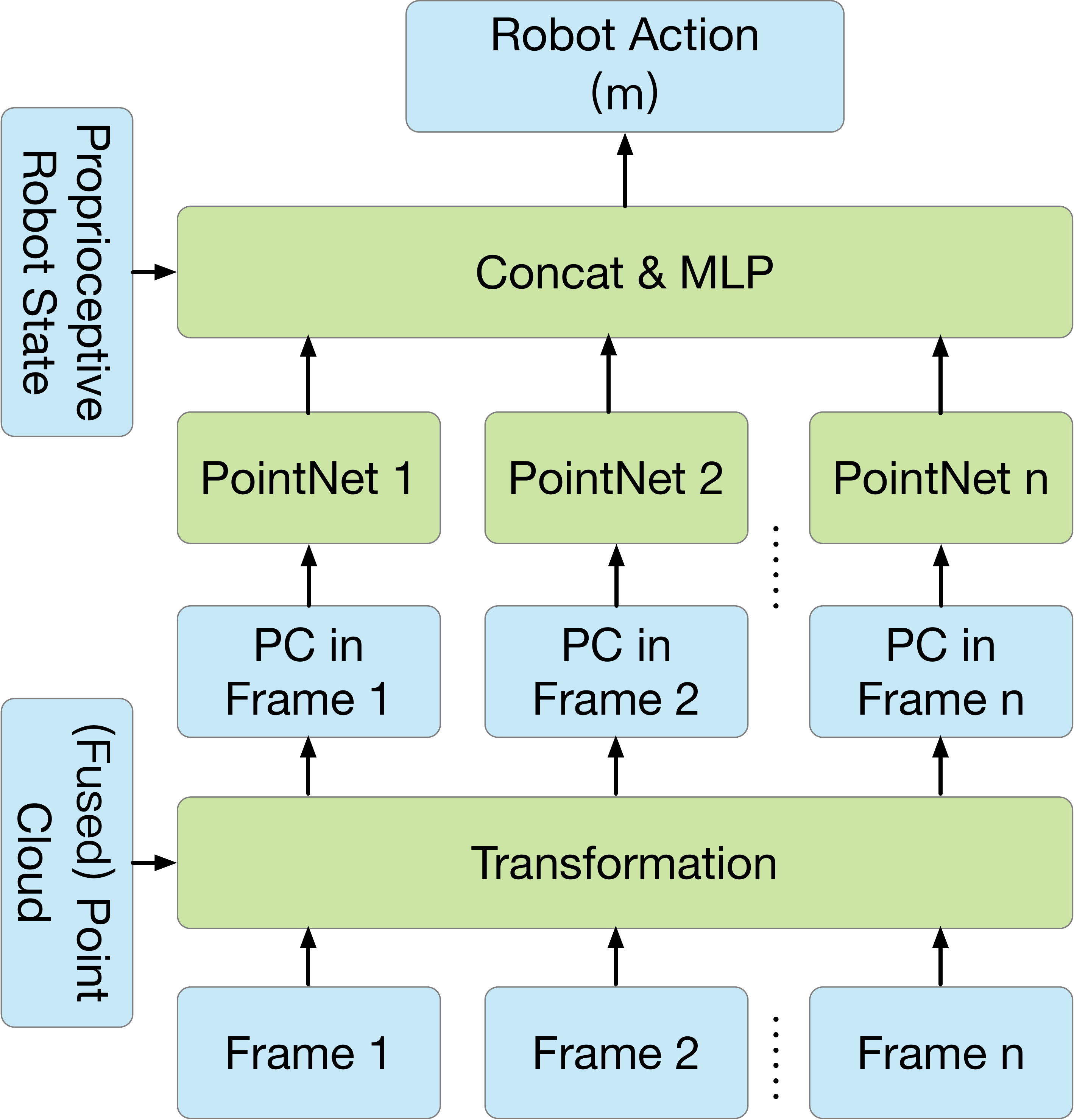}
         \caption{FrameMiner-FeatureConcat}
     \end{subfigure}
     \hfill
     \begin{subfigure}[t]{0.42\textwidth}
         \centering
         \includegraphics[width=\textwidth]{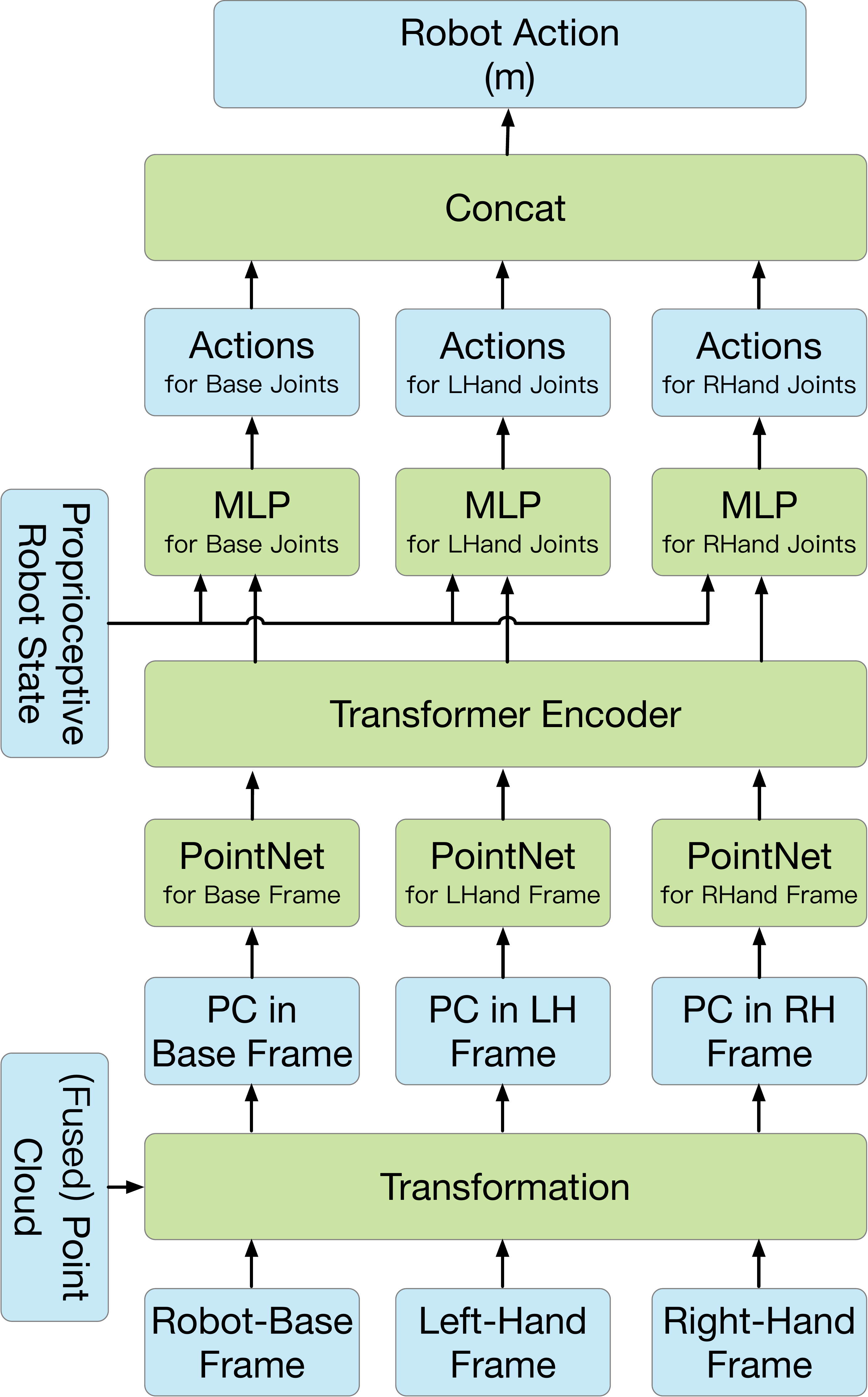}
         \caption{FrameMiner-TransformerGroup}
     \end{subfigure}
        \caption{Architectures of FrameMiner-FeatureConcat and FrameMiner-TransformerGroup.}
        \label{fig:FMFC,FMTG}
\end{figure}

\subsection{Additional Experiment Results and Discussions}

\subsubsection{Imitation Learning}
\label{sec:app_imitation}

In the main text, we analyzed the profound impact of coordinate frames on point cloud-based object manipulation learning through online RL algorithms. Apart from online RL, some previous work \citep{wen2022you} have shown that dynamic selection of coordinate frames could benefit demonstration-based manipulation learning as well. In this section, we conduct experiments on imitation learning and investigate whether our previous findings can generalize to other algorithm domains. 

For each task, we use an expert RL policy to generate 100 successful demonstrations. We then perform Behavior Cloning (BC) by representing input point clouds under different coordinate  frames, along with using our proposed FrameMiner-MixAction (FM-MA). We utilize the same network architectures as online RL, and we use MSE loss for training. For FM-MA, the robot-base frame and the end-effector frame(s) are fused. As shown in Table \ref{tab:imitation}, we observe similar findings to Section \ref{sec:single_results} and Section \ref{sec:multi_fusion_comparison}. Specifically, the end-effector frame has much higher performance on single-arm tasks (OpenCabinetDoor/Drawer), demonstrating the benefits of end-effector alignment. Our proposed FrameMiner is capable of automatically selecting the best single frame or combining the merits from multiple frames and outperforming single-frame baselines.
\begin{table}[t]
  \centering
  \setlength{\tabcolsep}{2.4pt}
    \begin{tabular}{c|ccc}
    \toprule
          & Robot-Base  & End-Effector   & FM-MA \\
    \midrule
    OpenCabinetDoor &  50$\pm$3    &  \textbf{85$\pm$3}     &  \textbf{83$\pm$4}      \\
    OpenCabinetDrawer &   72$\pm$4    &  \textbf{88$\pm$2}     &  \textbf{88$\pm$2}     \\
    PushChair &  38$\pm$3     &  28$\pm$2     &  \textbf{42$\pm$4}     \\
    MoveBucket &   76$\pm$4    &   80$\pm$2    &  \textbf{91$\pm$2}     \\
    \bottomrule
    \end{tabular}%
    \vspace{0.5em}
   \caption{Behavior Cloning (BC) success rates (\%) on four ManiSkill tasks. Mean and standard deviation over 5 seeds are shown.}
    \label{tab:imitation}
\end{table}

\subsubsection{Alternative Designs in FM-MA (Weighted Linear Combination vs. Maximum Weight)}
\label{sec:app_frameminer_max_weight}
\begin{figure}[t]
    \centering
    \includegraphics[width=0.5\textwidth]{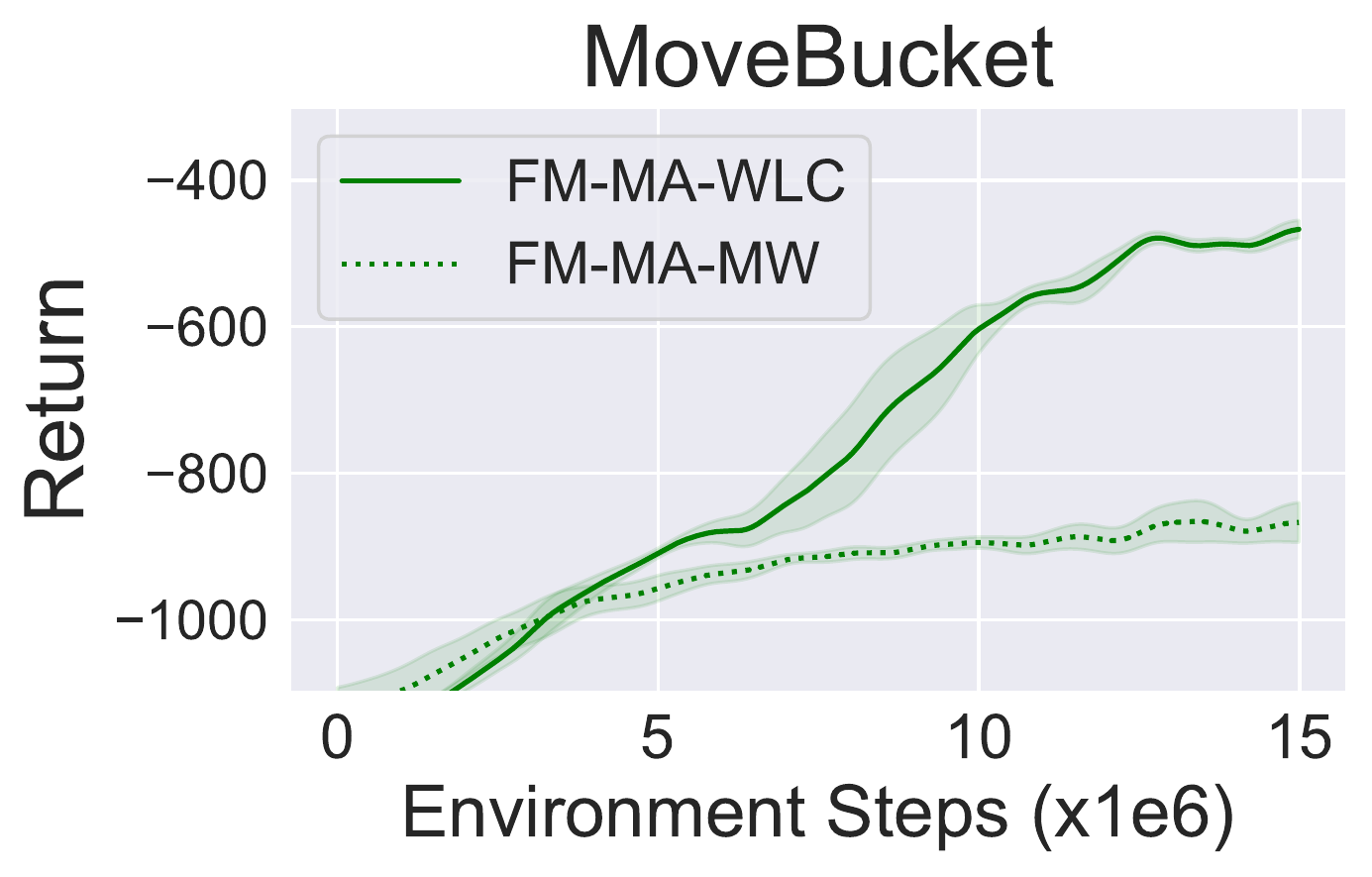}
    \caption{Comparison between FM-MA-WLC and FM-MA-MW on MoveBucket. Mean and standard deviation over 5 seeds are shown. FM-MA-WLC achieves 81$\pm$3\% final success rate, while FM-MA-MW only has 9$\pm$2\% final success rate. }
    \label{fig:max_action}
\end{figure}

\begin{table}[t]
  \centering
  \setlength{\tabcolsep}{2.4pt}
    \begin{tabular}{c|cc}
    \toprule
          & FM-MA (WLC eval) & FM-MA (MW eval) \\
    \midrule
    OpenCabinetDoor &  84$\pm$2    &  45$\pm$5           \\
    OpenCabinetDrawer &   93$\pm$1    &  93$\pm$2          \\
    PushChair &  36$\pm$4     &  20$\pm$3          \\
    MoveBucket &   81$\pm$3    &   14$\pm$3        \\
    \bottomrule
    \end{tabular}%
    \vspace{0.5em}
   \caption{Success rate (\%) comparison between the same FM-MA checkpoint evaluated using weighted linear combination of actions (WLC) and using maximum-weighted action (MW) on four ManiSkill tasks. Mean and standard deviation over 5 seeds are shown.}
    \label{tab:max_action}
\end{table}

In the main paper, FrameMiner-MixAction (FM-MA) uses weighted linear combination to fuse action proposals from each coordinate frame (see Figure \ref{fig:frame_miner}). For simplicity, we name this variant FM-MA-WLC. An alternative design is to choose the max-weighted action proposal for each joint (we name this variant FM-MA-MW). Formally, let $A \in \mathbb{R}^{n\times m}$, where $A_{ij}$ denotes the action proposal for the $j$-th robot joint from the $i$-th coordinate frame. Let $W \in \mathbb{R}^{n\times m}$ be the weight matrix predicted by the network. In FM-MA-MW, the output action $\mathbf{a}=(a_1,a_2,\dots,a_m)$ satisfies $a_j$ = $A_{kj}$ where $k=\textrm{argmax}_{k=1}^{n} W_{kj}$. Note that FM-MA-WLC uses SoftMax to normalize the weights; thus FM-MA-WLC can be regarded as a ``soft version'' of FM-MA-MW.

To compare the two designs, we conduct two experiments: 
(1) We train FM-MA-MW from scratch. Results are shown in Fig.~\ref{fig:max_action}. (2) We resume from the final checkpoint of the original FM-MA-WLC. During evaluation, we use the max-weighted action proposal as the action output. Results are shown in Table \ref{tab:max_action}. We observe that for both experiments, using FM-MA-MW deteriorates performance. We conjecture that FM-MA-WLC alleviates optimization difficulty, which likely comes from the fact that it is a ``soft version'' of FM-MA-MW with well-behaving gradients. On the other hand, since FM-MA-MW uses argmax operation over columns of $W$, there is a lack of gradient for $W$ during training, which leads to more difficult optimization.

\subsubsection{Ablation Study on Camera Placements}
\label{sec:app_cam_placement}
As a recap, the five tasks analyzed in our main paper cover both static and moving camera settings. The experiments in the main paper were conducted using default camera placements shown in Fig.~\ref{fig:tasks}. For the four tasks with moving cameras, a panoramic camera is mounted on the robot head.

\begin{figure}[t]
    \centering
    \includegraphics[width=0.5\textwidth]{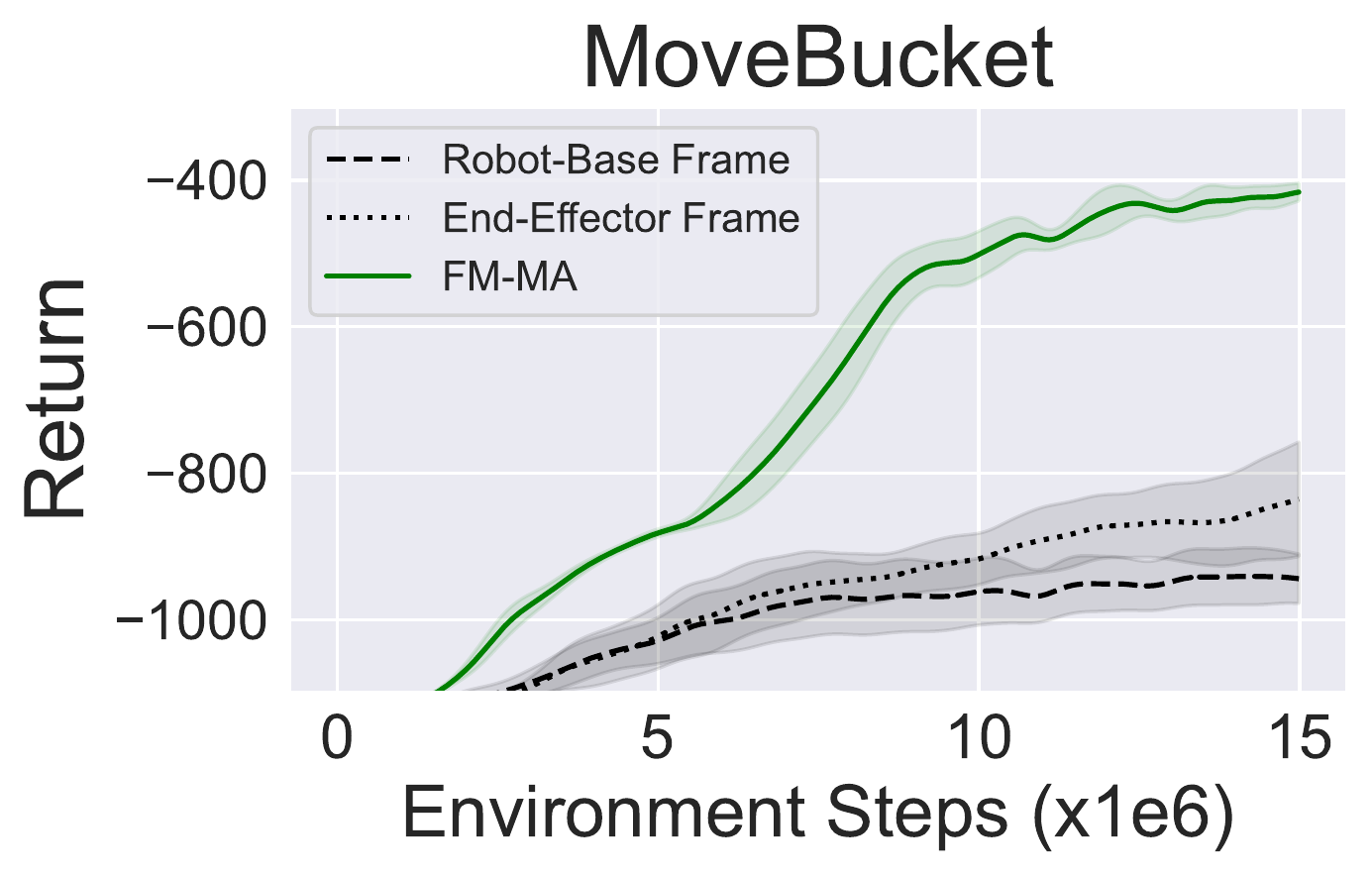}
    \caption{Results on MoveBucket with a panoramic camera mounted on the robot base. The ``Robot-Base Frame'' and the ``End-Effector Frame'' indicate the coordinate frames used to represent captured input point clouds. FM-MA fuses the two end-effector frames (left and right arms) and the robot-base frame. Mean and standard deviation over 5 seeds are shown.}
    \label{fig:cam_placements}
\end{figure}

While FrameMiners do not require changing existing camera placements, camera placements could still matter, since different camera placements affect the point clouds being captured (due to different occlusion and sparsity patterns). Therefore, we perform an experiment where we move the panoramic camera from the robot head to the robot base. As shown in Fig.~\ref{fig:cam_placements}, we observe similar phenomena as in Fig.~\ref{fig:multi_frame_comparison}. Specifically, fusing multiple coordinate frames with our FrameMiners still leads to better sample efficiency and final performance, demonstrating that FrameMiners are robust under different camera placements.

\subsubsection{Learning Adaptive Frame Transformations from Observations}
\label{sec:app_adaptive_se3}
\begin{figure}[t]
    \centering
    \includegraphics[width=0.5\textwidth]{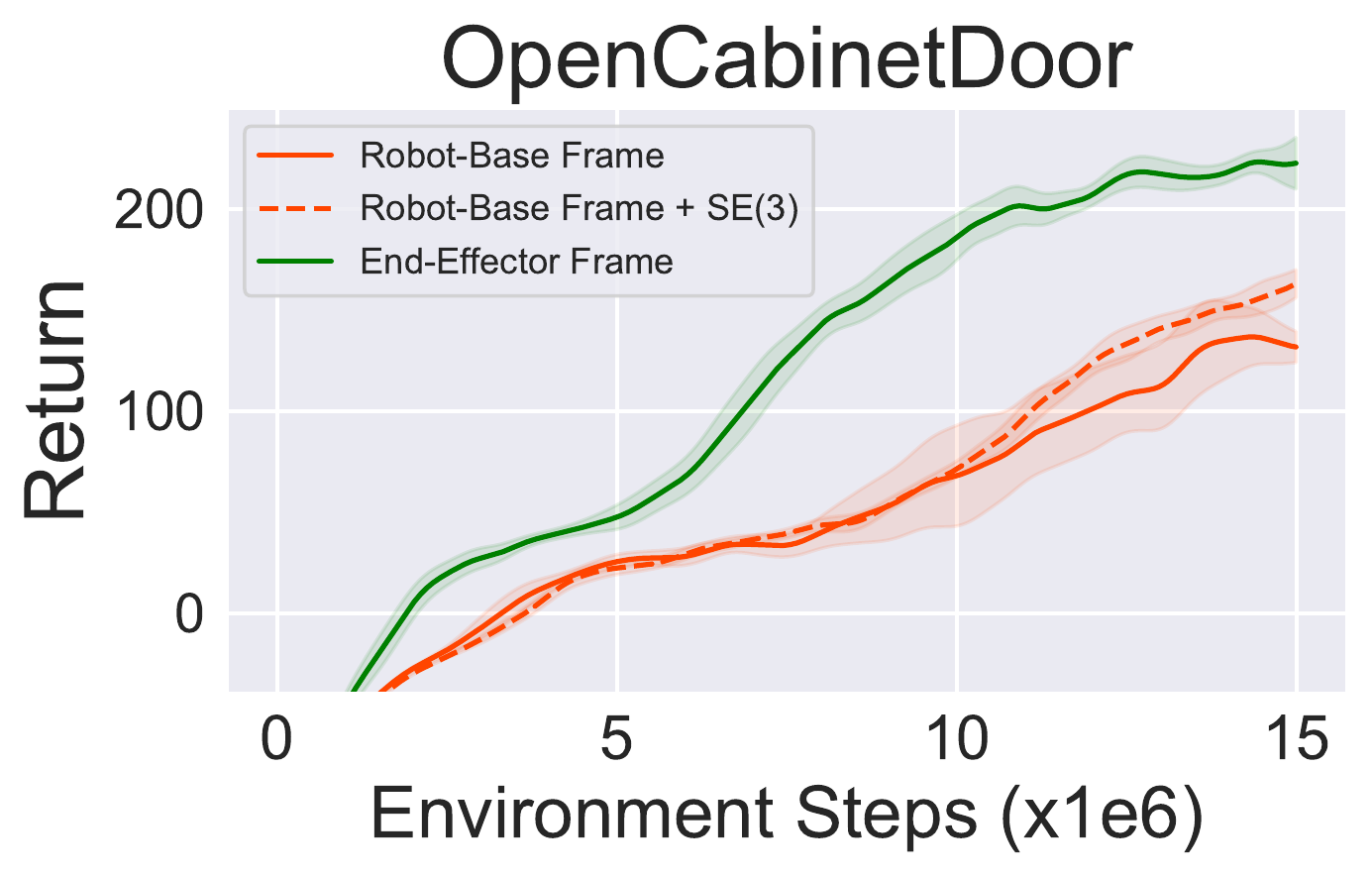}
    \caption{Ablation study on adding an adaptive $\mathbb{SE}(3)$ transformation prediction layer. When the input point cloud is represented in the robot-base frame, adding such transformation layer barely improves performance, while representing the point cloud in the end-effector frame significantly improves performance. }
    \label{fig:learnt_se3.pdf}
\end{figure}

In our paper, we use known transformations (e.g., end-effector pose in robot state) to align input point clouds in different coordinate frames and propose FrameMiners to fuse merits of multiple coordinate frames. A potential baseline is to learn a transformation adaptively based on input point clouds. To examine the effectiveness of this baseline, we add an additional network before the PointNet backbone to learn an adaptive $\mathbb{SE}(3)$ transformation based on the input point cloud. This transformation is then applied to the input point cloud before passing it through the PointNet backbone (note that we remove spatial transformation layers from the original PointNet in all of our experiments). However, as shown in Figure~\ref{fig:learnt_se3.pdf}, adding this $\mathbb{SE}(3)$ transformation layer barely improves performance. 

We conjecture that it's very difficult to predict a $\mathbb{SE}(3)$ transformation for aligning the input point cloud across time due to the large search space (where most transformations are ineffective) and weak supervision from RL training loss. Moreover, in many challenging tasks, we may need to fuse information simultaneously from multiple coordinate frames (e.g., left-hand and right-hand frames). This is not achievable through learning a single transformation. In contrast, for FrameMiners, we take advantage of easily-accessible frame information (e.g. end-effector poses) without relying on transformation prediction. We then fuse the merits of multiple candidate coordinate frames.

\subsubsection{$\mathbb{SO}(3)$ and $\mathbb{SE}(3)$ Equivariant Point Cloud Backbones}

Recently, there have been several works on designing $\mathbb{SO}(3)$ and $\mathbb{SE}(3)$ equivariant/invariant backbone networks for point cloud learning \citep{deng2021vector, simeonovdu2021ndf}. While they are of great benefit for analysis within each object (e.g., shape classification, part segmentation, and 6D pose estimation), our robot-object interaction setting is a bit different. 

In robot manipulation scenarios, a particular challenge comes from inferring the relations between two object parts (e.g., relative pose between the end-effector and the cabinet handle). This binary relation inference task is challenging under the weak RL loss supervision, even using $\mathbb{SO}(3)$ and $\mathbb{SE}(3)$ equivariant/invariant backbones. FrameMiners explicitly approach this challenge by aligning point clouds (across multiple time steps) with the known transformation matrices (e.g., the end-effector pose). This reduces many binary relation inference tasks to single-subject location tasks, which has much lower difficulty. For example, when using the end-effector frame in the OpenCabinet task, the network only needs to copy the handle pose to infer the relative pose between the handle and the end-effector, as the end-effector is always at the frame origin. 

\subsection{More Details of Manipulation Tasks}
\label{sec:app_task_descriptions}
\subsubsection*{Task Descriptions:} 
\begin{itemize}[leftmargin=*]
\setlength\itemsep{0em}

\item In OpenCabinetDoor, a single-arm mobile agent needs to approach a cabinet, use the handle to fully open the designated cabinet door, and then keep the door static for a while. 
\item In OpenCabinetDrawer, a single-arm mobile agent needs to approach a cabinet, use the handle to fully open the designated cabinet drawer, and then keep the drawer static for a while. 
\item In PushChair, a dual-arm mobile agent needs to approach the chair, push the chair to a target location, and then keep the chair static for a while. 
\item In MoveBucket, a dual-arm mobile agent needs to approach the bucket, move the bucket to a target platform, place the bucket onto the platform, and then keep the bucket static for a while. 
\item In PickObject, a single-arm fixed-base agent needs to grasp an object from the table, lift it up to a certain target height, and keep it static for a while. 
\end{itemize}

Simulations are fully physical. For OpenCabinetDoor, OpenCabinetDrawer, PushChair, and MoveBucket, there are 66, 49, 26, and 29 different objects (designated parts) during training, respectively.

\subsubsection*{Observations and Actions:} 

For all ManiSkill tasks, the proprioceptive robot state includes:
\begin{itemize}[leftmargin=*]
\setlength\itemsep{0em}
    \item Positions of all (two if single-arm, four if dual-arm) fingers
    \item Velocities of all (two or four) fingers
    \item x, y position of the mobile robot base
    \item Mobile robot base's rotation around the z-axis
    \item x, y velocity of the mobile robot base
    \item Angular velocity of the mobile robot base around the z-axis
    \item Joint angles of the robot,  excluding the joints in the mobile base
    \item Joint velocities of the robot, excluding the joints in the mobile base
    \item Indicator of whether each joint receives an external torque
\end{itemize}

The action space includes:

\begin{itemize}[leftmargin=*]
\setlength\itemsep{0em}
    \item x, y velocity of the mobile robot base
    \item Angular velocity of the mobile robot base around the z-axis
    \item Height of the robot body
    \item Joint velocities of the robot, excluding joints of the mobile base and the gripper fingers
    \item Joint positions of the gripper fingers
\end{itemize}

Joint positions of the gripper fingers are controlled by position PID. All other action components are controlled by velocity PID.

For the PickObject task, the proprioceptive robot state includes:
\begin{itemize}[leftmargin=*]
\setlength\itemsep{0em}
    \item Joint angles of the robot,
    \item Joint velocities of the robot, 
    \item 1D gripper joint position,
    \item Target xyz positions of object.
\end{itemize}

The action space includes 3 DoF end-effector position and 1 DoF gripper joint position.

For all tasks, input point cloud features include xyz coordinates, RGB colors, and one-hot segmentation masks for each part category.

\subsection*{Motivations for Our Task Choice}

We aim to cover a wide range of factors that may influence the selection of point cloud coordinate frames. Specifically, the tasks are chosen to cover various robot mobilities, numbers of robot arms, and camera settings, as demonstrated in Figure 1. 

Different robot mobility results in differences in world frame and robot base frame. These two frames are aligned in static robots but not in mobile robots. The robot's mobility can also change the focus of tasks (e.g.,  navigation or object interaction), which may place different requirements on the choice of point cloud frame.

We cover both single-arm and dual-arm environments, as they pose different requirements for point cloud frame selection. In single-arm environments, using the only end-effector frame may already be able to achieve good performance. However, in dual-arm environments, there are two end-effector frames, and these tasks require precise coordination between the two robot arms, which pose significant challenges for manipulation learning. As each end-effector may have a preferred frame, the necessity of frame fusion becomes more pronounced.

Last but not least, camera placements determine sources of point clouds, which may potentially influence the selection of coordinate frames. In our experiments, we cover both static camera settings and moving camera settings (mounted on robots). 

\begin{figure}[t]
    \centering
    \vspace{-1em}
    \includegraphics[width=0.5\textwidth]{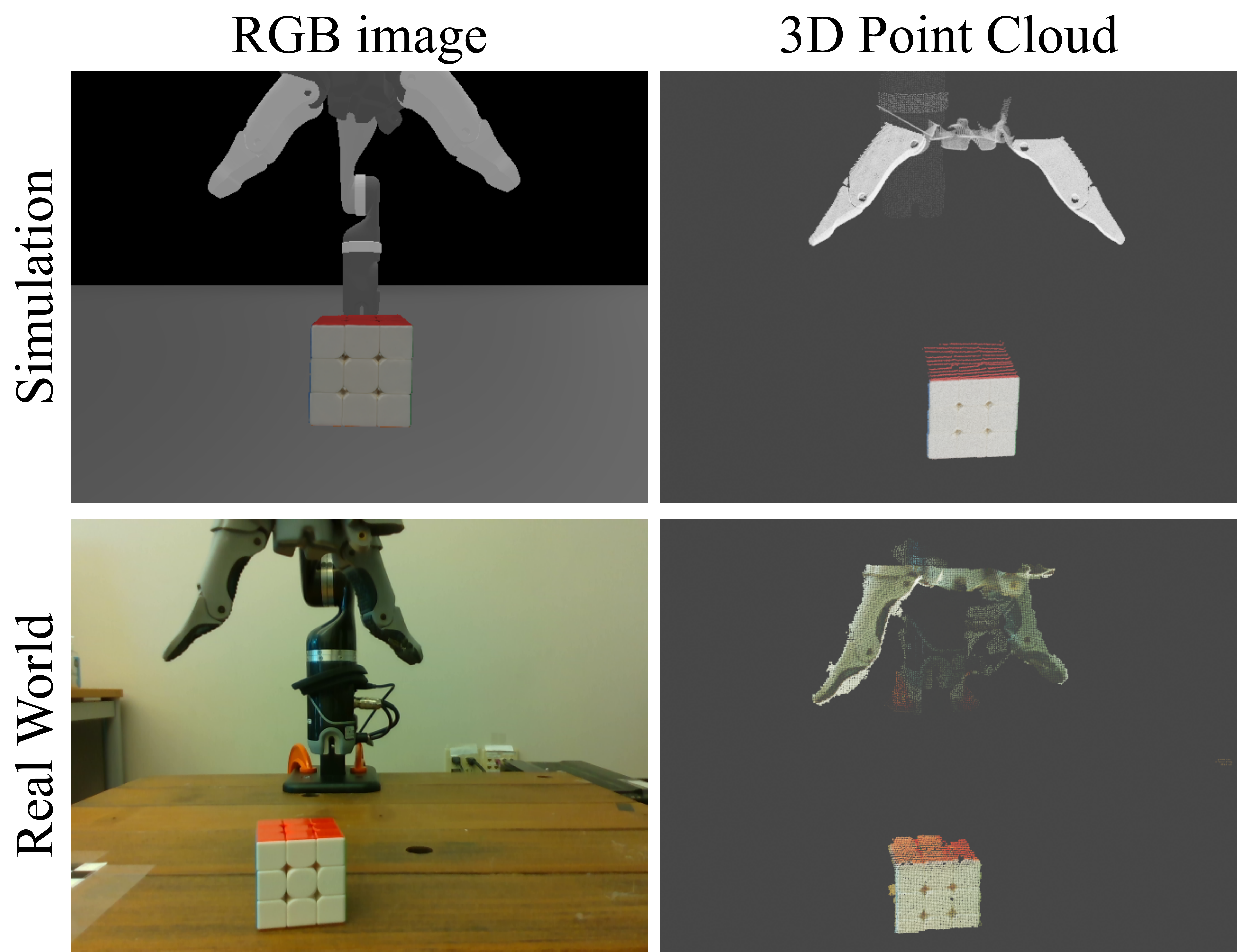}
    \caption{RGB images and 3D point clouds captured in both simulation and the real world. Colored point clouds for better illustration.}    
    \label{fig:real_world_supp}
    \vspace{-1em}
\end{figure}

\subsection{Detailed Experimental Settings and Hyperparameters}
\label{hyperparameters}
For our visual backbones, our PointNets are implemented with a three-layer MLP with dimensions $[64,128,300]$ followed by a max-pooling layer. We do not apply any spatial transformation to the inputs. Our SparseConvNets are implemented as a SparseResNet10 using TorchSparse~\citep{tang2020searching}. SparseResNet10 has a 4-stage pipeline with kernel size 3 and hidden channels $[64,128,256,512]$ respectively. We use kernel size 3 and stride 2 for downsampling. Initial voxel size is 0.05. Final features in the final-stage voxels are maxpooled as output visual feature.

All of our agents are trained with PPO (hyperparameters in Tab.~\ref{tab:ppo_hyper}).
Each policy MLP that outputs actions has dimensions $[192, 128, \textrm{action\_dim}]$. For FM-MA that uses input-dependent joint-specific weights to fuse action proposals from different frames, the MLP has dimension $[192, n\times m]$, where $n$ is the number of frames and $m$ is the dimension of action space. For FM-TG that uses Transformer to fuse features from different frames, the Transformer has 3 layers with hidden dimension 300 and feed-forward dimension 1024. For all network variants, the value head takes the concatenation of all visual features from all frames as input and passes through an MLP with dimensions $[192, 128, 1]$ to output value prediction. 

In addition, we found that zero-initializing the last layer of MLP before action output along with the joint-specific weights in FM-MA to be very helpful for stabilizing agent training.

For each task, we train an agent for a fixed number of environment steps. Specifically, for OpenCabinetDoor, OpenCabinetDrawer, and MoveBucket, we train for 15 million steps. For PushChair, we train for 20 million steps. For PickObject, we train for 4 million steps. Success rates are calculated among 300 evaluation trajectories.

\begin{table}[t]
\centering
\small
\begin{tabular}{r|cc}
\toprule
Hyperparameters & Value\\
\midrule
Optimizer & Adam \\
Discount ($\gamma$) & 0.95 \\
$\lambda$ in GAE & 0.95 \\
PPO clip range & 0.2 \\
Coefficient of the entropy loss term of PPO $c_{ent}$ & 0.0 \\
Advantage normalization & True \\
Reward normalization & True \\
Number of threads for collecting samples & 5\\
Number of samples per PPO update & 40000 \\
Number of epochs per PPO update & 2 \\
Number of samples per minibatch & 330 \\
Gradient norm clipping & 0.5 \\
Max KL & 0.2 \\
Policy learning rate & 3e-4 (non FM-TG); 1e-4 (FM-TG) \\
Value learning rate & 3e-4 \\
Action MLP Last Layer Initialization & Zero-init \\
\bottomrule
\end{tabular}
\vspace{1em}
\caption{Hyperparameters for PPO.}
\vspace{-1em}
\label{tab:ppo_hyper}
\end{table}

\subsection{More Details of Real-World Experiments}
\label{sec:app_real_world}
Fig.~\ref{fig:real_world_supp} shows the captured RGB images and point clouds in both simulation and the real world (by RealSense camera). For both simulation and the real-world environment, the ground points are removed using z-coordinate threshold or RANSAC, and the distant points are clipped.
To reduce the sim-to-real gap, we only use xyz coordinates as our input point cloud feature, and we discard RGB colors.